\documentclass[11pt]{article}
\usepackage{consilr2021}
\usepackage{times}
\usepackage{url}
\usepackage{latexsym}
\usepackage{verbatim}

\usepackage{tikzpeople}
\usepackage{cite}
\usepackage{amsmath,amssymb,amsfonts}
\usepackage{listings}
\usepackage{algorithmic}
\usepackage{float}
\usepackage{tikz}
\usepackage{graphicx}
\usepackage{textcomp}
\usepackage{xcolor}
\usepackage{url}
\usepackage{hyperref}
\usepackage[final]{pdfpages}
\usepackage{pgfplots}
\usepackage{caption}
\usepackage{subcaption}
\usetikzlibrary{shapes.callouts}
\usetikzlibrary{positioning,arrows,calc}
\usepackage[final]{pdfpages}

\tikzset{modal/.style={>=stealth',shorten >=1pt,shorten <=1pt,auto,node distance=1.5cm,semithick},world/.style={circle,draw,minimum size=0.5cm,fill=gray!15},point/.style={circle,draw,inner sep=0.5mm,fill=black},reflexive above/.style={->,loop,looseness=7,in=120,out=60},reflexive below/.style={->,loop,looseness=7,in=240,out=300},reflexive left/.style={->,loop,looseness=7,in=150,out=210},reflexive right/.style={->,loop,looseness=7,in=30,out=330}}

\def\BibTeX{{\rm B\kern-.05em{\sc i\kern-.025em b}\kern-.08em
    T\kern-.1667em\lower.7ex\hbox{E}\kern-.125emX}}
    
    \definecolor{lightGray}{gray}{0.9}
\newcommand\comppuzzlestotal{300}
\newcommand\comppuzzlesamb{50}
\newcommand\withoutquestionsall{3,056}
\newcommand\withoutquestionsent{1,130}
\newcommand\withoutquestionscont{1,926}

\newcommand\comppuzzleswamb{250}
\newcommand\withquestionsall{15,680}
\newcommand\withquestionsent{4,353}
\newcommand\withquestionscont{8,389}
\newcommand\withquestionsunknown{2,938}

\newcommand\knightpuzzles{300}
\newcommand\knightquestionsall{430}
\newcommand\knightquestionsent{215}
\newcommand\knightquestionscont{215}

\newcommand\knightpuzzleswamb{150}
\newcommand\kwithquestionsall{940}
\newcommand\kwithquestionsent{105}
\newcommand\kwithquestionscont{105}
\newcommand\kwithquestionsunknown{730}

\newcommand\zebrapuzzles{5}
\newcommand\zebraquestionsall{650}
\newcommand\zebraquestionsent{130}
\newcommand\zebraquestionscont{520}

\newcommand\ambzebrapuzzles{1}
\newcommand\ambzebraquestionsall{125}
\newcommand\ambzebraquestionsent{9}
\newcommand\ambzebraquestionscont{36}
\newcommand\ambzebraquestionsunknown{80}

\newcommand\total{\the\numexpr\comppuzzlesamb+\knightpuzzles+\zebrapuzzles\relax}
\newcommand\totalamb{\the\numexpr\comppuzzleswamb+\knightpuzzleswamb+\ambzebrapuzzles\relax}
\newcommand\totalq{\the\numexpr\withoutquestionsall+\knightquestionsall+\zebraquestionsall\relax}
\newcommand\totalqe{\the\numexpr\withoutquestionsent+\knightquestionsent+\zebraquestionsent\relax}
\newcommand\totalqc{\the\numexpr\withoutquestionscont+\knightquestionscont+\zebraquestionscont\relax}

\newcommand\puz{PuzzTE}
\newtheorem{puzzle}{Puzzle}

\title{A Puzzle-Based Dataset for Natural Language Inference}

\author{Roxana Szomiu and Adrian Groza\\
Department of Computer Science\\
Technical University of Cluj-Napoca, Romania\\
  {\tt Roxana.Szomiu@cs.utcluj.ro, Adrian.Groza@cs.utcluj.ro} \\}

\date{}

\begin{document}
\maketitle
\begin{abstract}
  We provide here a dataset for tasks related to natural language understanding and natural language inference.
The dataset contains logical puzzles in natural language from three domains: comparing puzzles, knighs and knaves, and zebra puzzles.
Each puzzle is associated with the entire set of atomic questions that can be generated based on the relations and individuals occurring in the text.
For each question we provide the correct answer: entailment, contradiction or ambiguity. The answer's correctness is verified against theorem provers.
Good puzzles have two properties: (i) each piece of information is necessary and (ii) no unnecessary information is provided. These properties make puzzles interesting candidates for machine comprehension tasks.

\end{abstract}

\section{Introduction}


Recognising textual entailment (RTE) is a classical problem in natural language processing (NLP), that aims to identify the relationship between sentence pairs, specifically if they entail or contradict each other, or neither of those.
RTE has a wide range of applications including question answering, spam detection, sentiment analysis. 

Datasets play an important role in assessing the RTE systems by evaluating their ability to predict the correct label-relationship for two sentences. Building a new dataset can be a difficult task because texts can be extremely rich source of information, but if they are unstructured, extracting information from them can be challenging and time-consuming. 

The identification of correct relation - entailment, contradiction, unknown - can be performed either through manual annotation or by automatic labeling. Our proposed dataset consists of various puzzle, and based on each puzzle's text we generate all possible questions and then using logical theorem provers, we provide the answers for each question.
The proposed dataset contributes in evaluating Question Answering applications. Most of the recent advances in question answering (QA) proposed deep learning solutions and the progress was made in open-domain question answering \newcite{minaee2021deep}.

Our contribution is the creation of dataset for textual entailment based on different types of logic puzzles. The dataset contains $\total$ unambiguous puzzles with 4,136 questions, and $\totalamb$  ambiguous puzzles with 16,745 questions. Starting from the puzzle's text, the generated FOL theory is taken by the Mace4/Prover9 to automatically assess the entailment between puzzle's text and the newly generated question.
Solving puzzles in first order logic can be a challenging task too.
To the best of our knowledge, there is no puzzle-based dataset designed to be used for Recognizing Textual Entailment, that contains the entire set of all possible questions, that can be answered from the initial text. In the absence of  a similar dataset, our approach based on Question-Answering technique, aims to generate question/answer pairs from the text of puzzles.

\section{Datasets for inference tasks}

This section briefly describes the available datasets that are widely used for textual entailment  or question answering task.

The Sentence Involving Compositional Knowledge (SICK) is large dataset of human intuitions on English sentences, collected through crowdsourcing by ~\newcite{marelli-etal-2014-sick}. 
SICK includes about 10,000 sentence pairs, each annotated for the degree of semantic relatedness and the type of entailment relation: entailment, contradiction, and neutral.
The entailment annotation led to 5,595 neutral pairs, 1,424 contradiction pairs, and 2,821 entailment pairs. The SICK dataset was constructed starting from two existing sets: the 8K ImageFlickr\footnote{\url{http://nlp.cs.illinois.edu/HockenmaierGroup/data.html}} and the  SemEval 2012 STS MSR-Video Description\footnote{\url{http://www.cs.york.ac.uk/semeval-2012/task6/index.php?id=data}}. 

The Excitement-Open-Platform (EOP) relies on various datasets such as RTE-3 English data set, Excitement dataset\footnote{\url{http://hlt-nlp.fbk.eu/technologies/textual-entailment-graph-dataset}} provided by ~\newcite{magnini-etal-2014-excitement,kotlerman-2015}, SICK\footnote{\url{https://zenodo.org/record/2787612\#.YF86KRKxU5k}}, described in ~\newcite{marelli-etal-2014-sick}, OMQ\footnote{\url{https://www.dfki.de/~neumann/resources/omqdata.html}}, which is a RTE-style dataset, semi-automatically created from manually categorized German customer requests. 
EOP consists of 800 English text-hypothesis (T-H) pairs for training and 800 T-H pairs for testing. 
Each pair is annotated with one of the three classes: entailment, non-entailment or unknown. Also, each pair is labeled with the appropriate text inference task: information extraction (IE), information retrieval (IR), question answering (QA), or summarization (SUM).

The Recognizing Textual Entailment\footnote{\url{https://tac.nist.gov//2011/RTE/index.html}} (RTE) datasets proposed by \newcite{rte-n} come from a series of textual entailment challenges and are are constructed based on news and Wikipedia text.
The PASCAL RTE datasets have been annotated for contradiction. They are marked for a 3-way decision in terms of entailment: "yes" (entails), "no" (contradicts), and "unknown" (doesn't entail but is not a contradiction). 
The datasets are not balanced since contradictions represent about 10\% of the data. 
 
The Guardian Headlines Entailment Training Dataset consists of around 32,000 pairs of sentences (16,233 for which entailment does hold and 16,249 for which it doesn't) automatically extracted from The Guardian using the provided API.
Since constructing manually entailment datasets is a time consuming task, so, they are fairly small, which means machine learning approaches perform sub-optimally. As Hickl et al. \newcite{Hickl05recognizingtextual} showed, automatically constructed datasets can improve the performance of systems using machine learning by up to ten percent. 

The Textual Entailment Graph was created within the Excitement project by ~\newcite{kotlerman-2015} as a gold standard data to evaluate the task of automatic Textual Entailment Graph (TEG) generation. 
The main difference between this task and the traditional RTE task is that the text pairs are not independent. The nodes in the graph are inter-connected via entailment edges, which should not represent contradicting decisions. For instance, if the edges $(u,v)$ and $(v,w)$  are in the graph, then the edge $(u,w)$ is implied by transitivity. The English dataset contains a text collection generated on the basis of 389 emails sent by customers of a railway company. Text fragments contain the customers feedback and are clustered into 29, for a total of 756 nodes and 7,830 edges.


~\newcite{williams-etal-2018-broad} have introduced the Multi-Genre Natural Language Inference (MNLI)\footnote{\url{https://www.nyu.edu/projects/bowman/multinli/}} containing 433k sentence pairs annotated with textual entailment labels. 
MNLI is based on the Stanford Natural Language Inference (SNLI) described by \newcite{bowman2015large}. 
MNLI covers a range of genres of spoken and written text, and supports a distinctive cross-genre generalization evaluation.
SNLI is a collection of 570k human-written English sentence pairs manually labeled for balanced classification with the labels entailment, contradiction, and neutral.

The SciTail dataset is the first entailment dataset created for the science question answering task by~\newcite{Khot_Sabharwal_Clark_2018}. 
Each premise-hypothesis pair is annotated  as entails or neutral.
About 43.3\% of the questions did not have a single supporting sentence, indicating that these questions either need multiple sentences for question answering or better retrieval results. From the remaining 56.7\% (i.e. 1,834 questions), the dataset contains 27,026 examples (10,101 with 'entails' label, 16,925 with neutral label) divided into train/dev/test splits with 23,596/1,304/2,126 examples. 
 

Wikipedia has been a source for building QA datasets.
\newcite{thorne2018fever} have proposed the Fact Extraction and VERification (FEVER), 
 a dataset that contains 185,445 claims manually verified against the sections from Wikipedia pages and classified as supported, refuted, or not-enough-info.
\newcite{yang-etal-2015-wikiqa} have developed the WikiQA dataset  It consists of a set of question-answer pairs, collected and annotated for open-domain question answer research. The corpus includes 3,047 questions and 29,258 sentences, where 1,473 sentences were labeled as answer sentences to their corresponding questions.
The dataset also includes questions for which there is no correct answer, allowing researchers to evaluate answer triggering models.
The Stanford Question Answering Dataset (SQuAD) proposed by \newcite{rajpurkar2016squad} is a collection of question-answer pairs extracted from Wikipedia articles. In SQuAD, the correct answers of questions can be any sequence of tokens in the given
text. The questions and answers are produced by crowdsourcing, so it is more diverse than some other question-answering datasets. 
The first version of SQuAD contains 107,785 question-answer pairs on 536 articles, while SQuAD2.0 combines the 100,000 questions in SQuAD1.1 with over 50,000 un-answerable questions written adversarially by crowdworkers in forms that are similar to the answerable ones.

\section{Question answering on puzzles}
Our puzzle based dataset for textual entailment (\puz) contains three types of puzzles: (i) comparison type puzzles, (ii) knight and knaves puzzles, and (iii) zebra puzzles.
The method used to build the dataset relies on natural language processing to automatically identify the characters from each puzzle and relationships among them. Based on manually defined grammar rules  we obtain a formal representation of each puzzle in First Order Logic (FOL). 
The puzzles were solved using Mace4 model finder and Prover9 theorem prover developed by ~\newcite{mccune2005prover9}. 
 We automatically generated all possible atomic questions against each puzzle and compute the correct answer based on automated deduction.
Hence, each QA pair is automatically annotated with three labels: (1) \textit{entailment} when a proof is found, (2)  $contradiction$ when the opposite sentence is proved, and (3) \textit{unknown} in case of not enough information in the puzzle, signaled by the Mace4 tool with more than one interpretation model.
Given specific grammar rules for each domain, the same automatic is applied for each type of puzzle exemplified here: comparison, knights and knaves, and zebra puzzles.

\subsection{Comparison type puzzles}
A comparison puzzle describes a scenario that involves ordering relationships. 
Take the example:
\begin{puzzle}
\textit{Mike is taller than Sally who is shorter than Katy. Ted is taller than Bob but shorter than Sally. Katy is shorter than Mike. Who is the tallest? Is Katy taller than Bob? Is Mike shorter than Ted? }
\label{puzzle:puzzle1}
\end{puzzle}


\begin{figure}
\begin{subfigure}[b]{0.32\textwidth}
\begin{center}
\begin{tikzpicture}
    \begin{scope}[every node/.style={thick,draw,scale=0.79},scale=0.79]
        \node (T) {Ted};
        \node[above of=T, xshift=7em,  yshift=2em] (K) {Katy};
        \node[right of=T, xshift=5em,  yshift=0em] (B) {Bob};
        \node[above of=T, xshift=-7em, yshift=2em] (S) {Sally};
        \node[above of=T, yshift=6em] (M) {Mike};
    \end{scope}
    
    \begin{scope}[>={Stealth[black]},every edge/.style={draw=black,thick}]
       \path [->] (M) edge node[sloped, above, font=\footnotesize]{taller} (S);
       \path [->] (S) edge node[sloped, above, font=\footnotesize]{shorter} (K);
       \path [->] (T) edge node[sloped, above, font=\footnotesize]{taller} (B);
       \path [->] (T) edge node[sloped, above, font=\footnotesize]{shorter} (S);
       \path [->] (K) edge node[sloped, above, font=\footnotesize]{shorter} (M);
    \end{scope}
\end{tikzpicture} 
\subcaption{Information extracted from Puzzle~\ref{puzzle:puzzle1}\label{fig:kk1p}}
\end{center}
\end{subfigure}
\begin{subfigure}[b]{0.71\textwidth}
\centering
\begin{small}
\begin{tabular}{l|l|l} \hline
Question & Theorem in FOL & Answer  \\ \hline
Is Mike the tallest ? & $tallest(mike)$ & Entailment \\
Is Bob the shortest ? & $shortest(bob)$ & Entailment \\
Is Katy taller than Bob ? & $taller(katy,bob)$ & Entailment \\
Is Bob shorter than Ted ? &  $shorter(bob,ted)$ & Entailment \\\hline
Is Sally the tallest ? & $tallest(sally)$ & Contradiction \\
Is Katy the shortest ? & $shortest(katy)$ & Contradiction \\
Is Bob taller than Katy ? & $taller(bob,katy)$ & Contradiction \\
Is Mike shorter than Ted? & $shorter(mike,ted)$ & Contradiction \\\hline
\end{tabular}
\end{small}
\subcaption{Sample of atomic question with 4 predicates and 5 individuals}
\label{tab:sample_comp}
\end{subfigure}

\end{figure}

By using the grammar rules, the Puzzle~\ref{puzzle:puzzle1} is automatically translated into FOL with three clauses: (1) $taller(Mike,Sally) \wedge  shorter(Sally,Katy)$, (2) $taller(Ted,Bob) \wedge shorter(Ted,Sally)$, respectively (3) $shorter(Katy,Mike)$. 
The manually defined grammar rules in NLTK (Listing~\ref{lst:grammarrules1}) are generic for all puzzles from this comparison puzzle domain.
The sequences of words are grouped into chunks. An example
of a chunk is the noun phrase chunk, which can be noun phrases such as "the girl" or "the boy", or proper names such as "Mike" or "Sally".
Unlike the words in lexical rules, which can be considered
terminals, the chunks are nonterminals. 
The linguistic categories have different properties (e.g. plural for nouns). One can specify specify the number for a noun with the feature \texttt{NUM}: \texttt{NP[NUM=sg,SEM=?subj]}. 
For verb phrases, the tense or person can be specified: \texttt{VP[NUM=sg, TNS=pres, PERS=third]}. 
These features help to assess the correctness of sentences. For instance, the context free grammar in Listing~\ref{lst:grammarrules1} can parse sentences like ”Mike is taller than Sally” or ”Mike is the tallest”, but won't be able to parse the sentence ”Katy are the tallest”. 

\begin{lstlisting}[caption=Sample of grammar rules for knights and knaves Puzzle~\ref{puzzle:puzzle1},firstline=3,lastline=15,linewidth=\columnwidth,breaklines=true,label=lst:grammarrules1]
S[SEM = <?subj(?vp)>] -> NP[NUM=?n,SEM=?subj] VP[NUM=?n,SEM=?vp]
NP[LOC=?l,NUM=?n,SEM=?np] -> PropN[LOC=?l,NUM=?n,SEM=?np] | InterPron[LOC=?l,NUM=?n,SEM=?np] | Nom[NUM=?n,SEM=?np] 
NP[+COP, NUM=?n, SEM=?nom ] -> Det[NUM=?n] Nom[NUM=?n, SEM=?nom]
NP[NUM=?n, SEM=<?det(?adj, ?nom)>] -> Det[NUM=sg, SEM=?det] Adj[GRD=?abs, SEM=?adj] Nom[NUM=sg,SEM=?nom] 
VP[NUM=?n,SEM=<?v(?prd)>] -> AuxP[+COP,NUM=?n,SEM=?v] Pred[SEM=?prd]
VP[NUM=?n,SEM=<?cc(?prd1, ?prd2)>] -> VP[NUM=?n,SEM=?prd1] CC[BOUND=?vpToVp, SEM=?cc] VP[NUM=?n,SEM=?prd2]
VP[NUM=?n,SEM=<?cc(?prd1, ?prd2)>] -> VP[NUM=?n,SEM=?prd1] CC[BOUND=?vpToVp, SEM=?cc] AdjP[GRD=?rel, SEM=?prd2]
AdjP[GRD=?rel, SEM=<?quant(?adj(?pp))>] -> Quantifier[SEM=?quant] Adj[GRD=?rel, SEM=?adj] PP[+than,SEM=?pp]
AdjP[GRD=?rel, SEM=<?adj(?pp)>] -> Adj[GRD=?rel, SEM=?adj] PP[+than,SEM=?pp]
AdjP[GRD=?rel, SEM=<?neg(?adj, ?pp)>] ->  Neg[SEM=?neg] Adj[GRD=?rel, SEM=?adj] PP[+than,SEM=?pp]
AdjP[GRD=?abs, SEM=<?adj>] -> Det[NUM=sg] Adj[GRD=?abs, SEM=?adj] 
PP[+than, SEM=?np] -> P[+than] NP[SEM=?np]
AuxP[COP=?c,NUM=?n,SEM=<?neg(?aux)>] -> Aux[COP=?c,NUM=?n,SEM=?aux] Neg[SEM=?neg]
\end{lstlisting}

The lexical rules in Listing~\ref{lst:grammarrules2} contain the linguistic categories for each puzzle. 
Each word belongs to such a category and can have one or more features (\newcite{Wagner2010StevenBE}). 
Two such features used by the discourse checker module are \texttt{SEM}, which specifies the semantic of the word and \texttt{NUM} which defines the number of a noun (sg-singular or pl-plural). 
We used the \texttt{GRD} feature that handle the gradable adjectives (e.g. relative gradable adjective  \textit{shorter} or \textit{taller}, or absolute gradable adjective (e.g. \textit{shortest}, \textit{tallest}).

\begin{lstlisting}[caption=Lexical rules for Puzzle~\ref{puzzle:puzzle1},firstline=18,lastline=35,linewidth=\columnwidth,breaklines=true,label=lst:grammarrules2]
PropN[-LOC,NUM=sg,SEM=<\P.P(Mike)>] -> 'Mike'
PropN[-LOC,NUM=sg,SEM=<\P.P(Sally)>] -> 'Sally'
PropN[-LOC,NUM=sg,SEM=<\P.P(Katy)>] -> 'Katy' 
PropN[-LOC,NUM=sg,SEM=<\P.P(Ted)>] -> 'Ted'
PropN[-LOC,NUM=sg,SEM=<\P.P(Bob)>] -> 'Bob'
Det[NUM=sg,SEM=<\P Q x.(P(x) & Q(x))>] -> 'the'
InterPron[-LOC, NUM=sg, SEM=<\P.P(x)>] -> 'who' | 'Who'
RelPron[SEM=<\Q x P.((P(x) & Q(x)))>] -> 'who'
Aux[+COP,NUM=sg,SEM=<\P x.P(x)>,tns=pres] -> 'is' | 'Is'
Aux[+COP,NUM=,SEM=<\P x.P(x)>,tns=pres] -> 'are' | 'Are'
CC[BOUND=?vpToVp, SEM=<\P Q x.((P(x) & Q(x)))>] -> 'but'
Adj[GRD=rel, SEM=<\X x.X(\y.(taller(x,y)))>] -> 'taller'
Adj[GRD=rel, SEM=<\X x.X(\y.(shorter(x,y)))>] -> 'shorter'
Adj[GRD=abs, SEM=<\x.(shortest(x))>] -> 'shortest'
Adj[GRD=abs, SEM=<\x.(tallest(x))>] -> 'tallest'
Quantifier[SEM=<\P. P>] -> 'little'
Quantifier[SEM=<\P. P>] -> 'too'
Question[] -> '?'
\end{lstlisting}

\begin{figure}
    \centering
    \includegraphics[width=16cm]{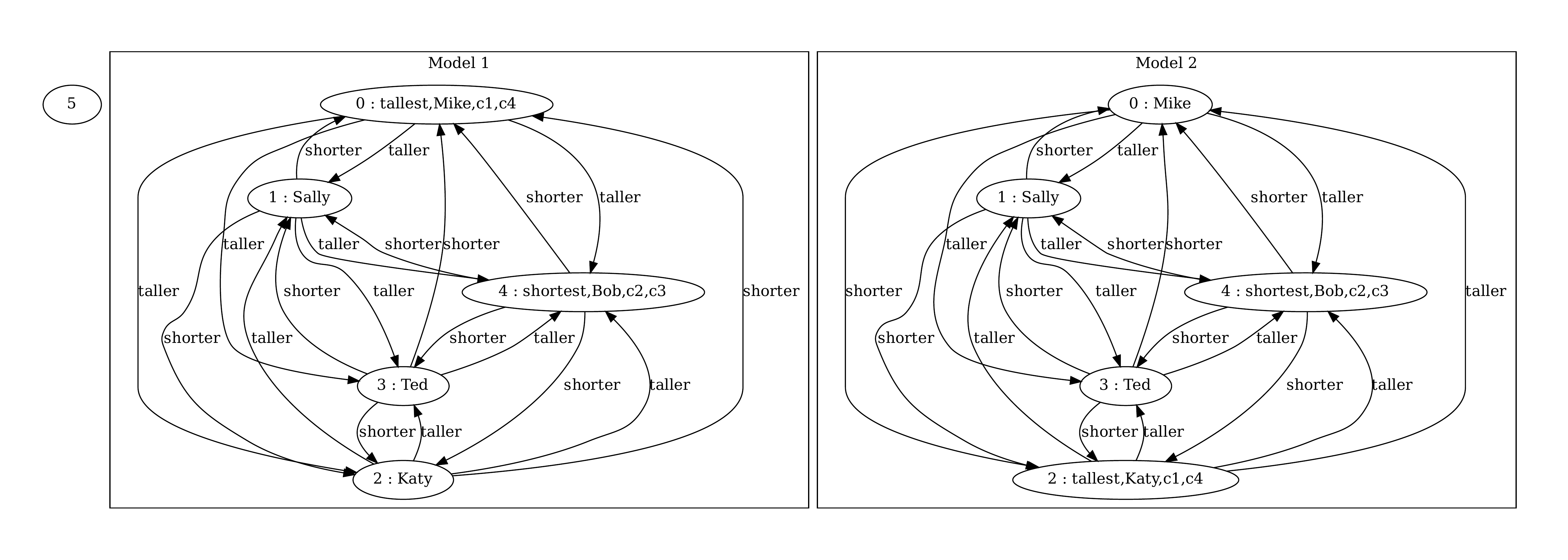}
    \caption{Models computed by Mace4 for Puzzle~\ref{puzzle:puzzle1} and Puzzle~\ref{puzzle:puzzle2}\label{fig:macemodels}}
\end{figure}

Based on the obtained FOL theory, Mace4 computes the models satisfying the given relation. 
For Puzzle~\ref{puzzle:puzzle1}, Mace4 generates a single model - the first one from Figure~\ref{fig:macemodels}.
By querying this model we can automatically  generate entailment or non-entailments answers. 
The set of atomic questions is generated based on the number of individuals from puzzle and the relations between. 
For the Puzzle~\ref{puzzle:puzzle1}, there are two unary predicates: \textit{shortest(x)} and \textit{tallest(x)} and two binary predicates: \textit{taller(x,y)} and \textit{shorter(x,y)}, and 5 individuals: \textit{Mike, Sally, Katy, Bob and Ted}. 
For each puzzle we generated a number of $a*n + b*n^2$ \label{formula:f1} questions, where $a$ is the number of unary predicates, $b$ the number of binary predicates, and $n$ the number of individuals in the given puzzle.
For this puzzle, we generated $2*5 +2 *5^2= 60$ questions.
Among them 22 will generate entailments, and 38 will generate contradictions. A sample of these 60 questions appears in Table~\ref{tab:sample_comp}.

To extend the dataset with more questions, two approaches can be used: (1) to consider other predicates based on some background knowledge, and (2) to generate puzzles with ambiguity by removing a part of information from the initial puzzle.

First, additional predicates can be obtained from some domain knowledge base. For the comparison puzzle domain one binary predicate can be \textit{sameTallAs(x,y)} or \textit{sameShortAs(x,y)}. 
Some additional questions could be "Are Katy as tall as Mike?" (i.e. $sameTallAs(Katy, Mike)$) or "Are Ted not as short as Bob?" (i.e. $\neg sameShortAs(Ted, Bob)$). 

Second, we introduced ambiguity by removing sentences from puzzles. 
We used the term "ambiguous" for puzzles that are not completely defined: the case when a part of information is missing that prevents us to find a single model. 
For ambiguous puzzles,  computes several interpretations which are models of the input formulas.
To generate ambiguous puzzles, we remove sentences (i.e. clues) from the complete puzzles. 
For instance, by removing \textit{Katy is shorter than Mike} from Puzzle~\ref{puzzle:puzzle1}, we obtain the ambiguous Puzzle~\ref{puzzle:puzzle2}.

\begin{puzzle}
\textit{Mike is taller than Sally who is shorter than Katy. Ted is taller than Bob but shorter than Sally. Who is the tallest? Is Katy taller than Bob? Is Mike shorter than Ted? }
\label{puzzle:puzzle2}
\end{puzzle}

Because of the missing information, one cannot infer whether \textit{Katy is shorter than Mike } or  \textit{if Katy is the tallest}. 
Mace4 signals this ambiguity by computing two models for Puzzle~\ref{puzzle:puzzle2} presented in Figure~\ref{fig:macemodels}.

We quantify the ambiguity of each puzzle by the number of \textit{unknown} answers. 
For Puzzle~\ref{puzzle:puzzle2}, the generated FOL theory contains two unary predicates: \textit{shortest(x)} and \textit{tallest(x)}, two binary predicates: \textit{taller(x,y)} and \textit{shorter(x,y)}, and 5 individuals: \textit{Mike, Sally, Katy, Bob and Ted}. Note that by removing some sentences, some predicates or even individuals may disappear from the puzzle.
Among all 60 questions, there are now 19 entailment answers, 35 contradictions, and 6 unknown answers. A sample of these 60 questions are listed in Table~\ref{tab:sample_comp2}. 

\begin{figure}
\begin{subfigure}[b]{0.65\textwidth}
\centering
\begin{small}
\begin{tabular}{l|l|l} \hline
Question & Theorem in FOL & Answer  \\ \hline
Is Katy taller than Sally ? & $taller(Katy,Sally)$ & Entailment \\
Is Mike taller than Sally ? & $taller(Mike,Sally)$ & Entailment \\
Is Mike taller than Katy ? & $taller(Mike,Katy)$ & Unknown \\
Is Katy taller than Mike ? & $taller(Katy,Mike)$ & Unknown  \\
Is Katy the tallest ? & $tallest(Katy)$ & Unknown \\
Is Mike the tallest ? & $tallest(Mike)$ & Unknown \\
Is Sally taller than Katy ? & $taller(Sally,Katy)$ & Contradiction \\
Is Mike shorter than Ted? & $shorter(mike,ted)$ & Contradiction \\\hline
\end{tabular}
\end{small}
\subcaption{Sample of atomic questions for ambiguous puzzles}
\label{tab:sample_comp2}
\end{subfigure}
\begin{subfigure}[b]{0.3\textwidth}
\begin{center}
  \begin{tikzpicture}[font=\small]
    \begin{axis}[
      width=7cm, 
      height=5cm,
      ybar,
      bar width=15pt,
      ymin=0,
      xtick=data,
      axis x line=bottom,
      axis y line=left,
      enlarge x limits=0.2,
      symbolic x coords={$< 15\%$, $15-25\%$, $25-50\%$, $> 50\%$},
      xticklabel style={
         anchor=base,
         yshift=-\baselineskip,
         text width=1.2cm, 
         align=center, 
      },
      nodes near coords={\pgfmathprintnumber\pgfplotspointmeta}
    ]

      \addplot[fill=orange,fill opacity=0.5, text opacity=1] coordinates {
        ($< 15\%$, 112)
        ($15-25\%$, 72)
        ($25-50\%$, 60)
        ($> 50\%$, 6)
      };
  \end{axis}
  \end{tikzpicture}
\subcaption{Distribution of ambiguity \label{fig:hist}}
\end{center}
\end{subfigure}
\end{figure}

By removing one statement or two statements from each puzzle, we can play with the level of ambiguity within the \puz\ dataset. The level of ambiguity is affected also by the information containing in each sentence. For instance the statement "Mike is taller than Sally who is shorter than Katy" contains more information than "Katy is shorter than Mike".  
Hence, the \puz\ dataset contains puzzles with different levels of ambiguity. The picture \ref{fig:hist} shows a histogram with the distribution of ambiguity in puzzles: there are 112 puzzles that have level of ambiguity less than 15 \%, there are 72 puzzles with level of ambiguity between 15-25 \%,  there are 60 puzzles with level of ambiguity between 25-50 \%,  and only 6 puzzles which have level of ambiguity greater than 50\%.

The \puz~dataset contains \comppuzzlestotal\  comparison type puzzles: \comppuzzlesamb\ puzzles completely described (without any ambiguity), and \comppuzzleswamb\ puzzles with different level of ambiguity. Part of these puzzles are created by as, and part of them were collected from online sources, mentioned in the dataset.\footnote{The PuzzTe dataset is available on Kaggle at this \href{https://www.kaggle.com/dataset/783dd6e298bfdc1d761d31fe5818bdefa9b2acc8c2879bc8b753f3966a177c6c}{Link}}

\subsection{Knights and knaves puzzles}

The second part of our dataset is built on  \knightpuzzles\ puzzles with knights and knaves taken from \url{https://philosophy.hku.hk/think/logic/knights.php}.
The complexity of these puzzles depends on the number of the individuals (ranging from 2 to 9) and on the type of sentences to be translated into FOL. Consider the Puzzle \ref{puzzle:puzzle3}) with 4 inhabitants: 

\begin{puzzle}
\textit{On the island where each inhabitant is either a knave or a knight , knights always tell the truth while knaves always lie. You meet four inhabitants: Bart, Dave, Rex and Sue. Bart tells you that Rex and Dave are both knights or both knaves. Dave says that Sue is a knave. Rex claims that Bart is a knave. Sue claims that Rex is a knight and Dave is a knave. Who is a knight and who is a knave?}
\label{puzzle:puzzle3}
\end{puzzle}

\begin{table}
\centering
\begin{small}
\begin{tabular}{l|l|l}\hline
Question & Theorem in FOL & Answer \\ \hline
Sue is a knight & $knight(Sue)$ & Entailment\\
Bart is a knave & $knave(Bart)$ & Entailment\\
Rex is a knight & $knight(Rex)$ & Entailment\\
Dave is a knave & $knave(Dave)$ & Entailment\\\hline
Sue is a knave & $knave(Sue)$ & Contradiction\\
Bart is a knight & $knight(Bart)$ & Contradiction\\
Rex is a knave & $knave(Rex)$ & Contradiction\\
Dave is a knight & $knight(Dave)$ & Contradiction\\ \hline
\end{tabular}
\end{small}
\caption{Question answering for Knights and Knaves puzzles\label{tab:qa_kk}}
\end{table}




Firstly, we formalise the common knowledge in all knights and knaves puzzles. The sentence \textit{"The inhabitants are either knights or knaves"} can be translated into FOL with $\forall x\ (inhabitant(x) \rightarrow knight(x) \vee knave(x))$. 
The sentence \textit{"One cannot be a knight and a knave in the same time"} is formalised with
 $\forall x\ (knight(x) \leftrightarrow \neg knave(x))$. 
A message $m(x)$ conveyed by a knight $x$ is always true: $knight(x) \rightarrow  m(x)$.
Similarly, a message $m(x)$ conveyed by a knave $x$ is always false: $knave(x) \rightarrow  \neg m(x)$.
This background knowledge appears in Listing \ref{lst:kk0bp}.

Second, we automatically translate the puzzle into a FOL theory.
We learn that there are four inhabitants, $Sue$, $Bart$, $Rex$ and $Dave$: $inhabitant(Sue) \wedge inhabitant(Bart) \wedge inhabitant(Rex) \wedge inhabitant(Dave)$, 
Hence, we compute a domain size of four.
The message of $Sue$ is: $m(Sue) \leftrightarrow  knight(Rex) \wedge knave(Dave)$, the message of $Dave$ is: $m(Dave) \leftrightarrow  knave(Sue)$,  the message of $Rex$ is: $m(Rex) \leftrightarrow  knave(Bart)$, while the message of Bart is:  $m(Bart) \leftrightarrow (knight(Rex) \wedge knight(Dave)) \vee (knave(Rex) \wedge knave(Dave)) $.

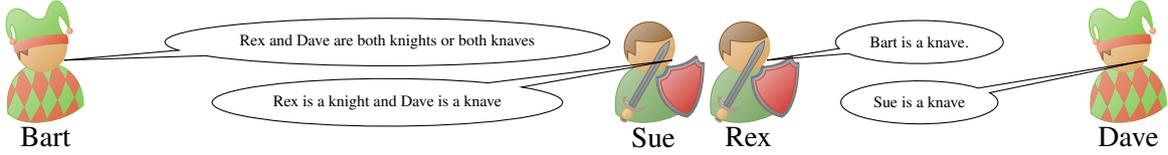
\begin{figure}
\begin{center}
\begin{tikzpicture}[scale=1,every node/.style={transform shape},roundnode/.style={circle, draw=green!60, fill=green!5, very thick, minimum size=7mm},]
\node[name=a,shape=jester,minimum size=1cm,xshift=-4.5cm] {Bart};
\node[name=b,shape=person,shield,sword, minimum size=1cm,xshift=3.5cm] {Sue};
\node[name=c,shape=person,shield,sword, minimum size=1cm,xshift=4.75cm] {Rex};
\node[name=d,shape=jester, minimum size=1cm,xshift=9.75cm] {Dave};
\node[ellipse callout, draw, yshift= .4cm,  callout absolute pointer={(a.mouth)}, font=\tiny] {Rex and Dave are both knights or both knaves};
\node[ellipse callout, draw, yshift=-.4cm, callout absolute pointer={(b.mouth)}, font=\tiny] {Rex is a knight and Dave is a knave};
\node[ellipse callout, draw, yshift= .4cm, xshift =7cm, callout absolute pointer={(c.mouth)}, font=\tiny] {Bart is a knave.};
\node[ellipse callout, draw, yshift=-.4cm, xshift =7cm, callout absolute pointer={(d.mouth)}, font=\tiny] {Sue is a knave};

\end{tikzpicture}
\caption{The single model computed by Mace4 for Puzzle~\ref{puzzle:puzzle3}\label{fig:kk0p}}
\end{center}
\end{figure}

To obtain the above theory, we formalise the grammar rules depicted in Listing~\ref{lst:grammarrules2}, that specify how words from different parts of speech are connected . 
We rely on the \texttt{SEM} feature, that defines the semantic of the words based on the lambda operator. For instance,  \textbackslash$x.(inhabitant(x) \& knight(x))$ represents the elements from the domain that are both inhabitant and knight too. The rule for compound sentences (line 5) parses statements like \textit{"Bart tells you that Rex and Dave are both knights or both knaves"}.
Given to Mace4, it computes the single model from Figure~\ref{fig:kk0p}.

\begin{lstlisting}[caption=Grammar rules for Puzzle~\ref{puzzle:puzzle3},firstline=41,lastline=50,linewidth=\columnwidth,breaklines=true,label=lst:grammarrules1]
S[SEM=<?np(?vp)>] -> NP[NUM=?n, SEM=?np] VP[NUM=?n, SEM=?vp]
S[SEM=?vp] -> Pron[NUM=?n] VP[NUM=?n2, SEM=?vp]
S[SEM=<?that(?sent1, ?sent2)>] -> S[SEM=?sent1] Conj[BOUND=sentences, TYPE=thatClause, SEM=?that] CompoundSent[SEM=?sent2]
CompoundSent[SEM=<?conjEither(?sent1, ?sent2)>] -> Conj[+either,SEM=?conjEither] S[SEM=?sent1] Conj[BOUND=sentences, SEM=?conj2] S[SEM=?sent2]
CompoundSent[SEM=<?conj2(?sent1, ?sent2)>] -> Quantifier[+BOTH] S[SEM=?sent1] Conj[BOUND=sentences, SEM=?conj2] S[SEM=?sent2]
PP[TYPE=?type] -> P[+on] NP[TYPE=?type, NUM=?n]
NP[TYPE=where, NUM=?n] -> Det[SEM=?det] Nom[NUM=?n, SEM=?nom]
NP[NUM=?n, SEM=?nom] -> Nom[NUM=?n, SEM=?nom]
NP[LOC=?l,NUM=?n,SEM=?np] -> PropN[LOC=?l,NUM=?n,SEM=?np] 
NP[NUM=?n, SEM=?nom] -> Numeral[NUM=?n] Nom[NUM=?n, SEM=?nom]
\end{lstlisting}



\begin{lstlisting}[caption=Reusing knowledge for the knight and knaves puzzles, firstline=2,lastline=4, linewidth=\columnwidth,breaklines=true,label=lst:kk0bp]
  all x (inhabitant(x) -> knight(x) | knave(x)). 
  all x ((knight(x) -> -knave(x)) & (knave(x) -> -knight(x))).
  knight(x) ->  m(x).                      knave(x)  -> -m(x).
\end{lstlisting}

\begin{lstlisting}[caption=Generating the FOL theory for Puzzle~\ref{puzzle:puzzle3},firstline=6,lastline=11,linewidth=\columnwidth,breaklines=true,label=lst:kk0p]
  inhabitant(Sue) & inhabitant(Bart) & inhabitant(Rex) & inhabitant(Dave).
  m(Bart) <-> (knight(Rex) & knight(Dave)) | (knave(Rex) & knave(Dave)). 
  m(Dave) <->  knave(Sue). 
  m(Rex)  <->  knave(Bart). 
  m(Sue)  <->  knight(Rex) & knave(Dave).
  Sue!=Rex & Sue!=Bart & Sue!=Dave & Rex!=Bart & Rex!=Dave & Bart!=Dave.
\end{lstlisting}

The resulted theory contains two unary predicates: \textit{knight(x)} and \textit{knave(x)} and four individuals: \textit{Bart, Sue, Dave and Rex}. Hence we generate $2 \times 4 = 8$ questions. 
Among them, 4 will generate entailments, and 4 will generate contradictions. A sample of these 8 questions are listed in Table~\ref{tab:qa_kk}.
The generated formalisation in FOL appears in Listings~\ref{lst:kk0bp} and \ref{lst:kk0p}. Note that the knowledge is divided into two modules: the background knowledge for the knight and knaves puzzles, that is reused for all puzzles in the domain, and the specific information extracted from Puzzle~\ref{puzzle:puzzle3}. 

To extend the dataset with unknown questions, we generate ambiguous puzzles by removing a part of the information. 
For instance, by removing \textit{Sue claims that Rex is a knight and Dave is a knave} from Puzzle~\ref{puzzle:puzzle3},  Mace 4 will generate two different models: in the first one, \textit{Dave and Rex are knaves and Bart and Sue are knights}, and in the second model: \textit{Rex and Sue are knights and Bart and Dave are knaves}.
For this ambiguous puzzles, for 2 questions Mace4 generates entailments,  for 2 questions Mace4 generates contradictions, and there are 4 questions that have unknown result (i.e Mace4 generates models in which that statements are both true or false (Table~\ref{tab:qa_kk2}).

\begin{table} 
\centering
\begin{small}
\begin{tabular}{l|l|l}\hline
Question & Theorem in FOL & Answer \\ \hline
Sue is a knight & $knight(Sue)$ & Entailment\\
Dave is a knave & $knave(Dave)$ & Entailment\\
Bart is a knave & $knave(Bart)$ & Unknown\\
Rex is a knight & $knight(Rex)$ & Unknown\\
Rex is a knave & $knave(Rex)$ & Unknown\\
Bart is a knight & $knight(Bart)$ & Unknown\\
Dave is a knight & $knight(Dave)$ & Contradiction\\
Sue is a knave & $knave(Sue)$ & Contradiction\\ \hline
\end{tabular}
\end{small}
\caption{Generating answers from ambiguous Knights and Knaves puzzles\label{tab:qa_kk2}}
\label{tab:sample_comp5}
\end{table}

\subsection{Zebra puzzles}
The third part of the \puz\ dataset includes zebra puzzles, whose  
complexity depends on the number of the individuals or number of houses and on the type of hints/clues to be translated into FOL. 

\begin{puzzle}
\textit{There are 5 houses in five different colors. In each house lives a person with a different nationality. These five owners drink a certain type of beverage, smoke a certain brand of cigar and keep a certain pet. No owners have the same pet, smoke the same brand of cigar or drink the same beverage. The question is: Who owns the fish?} 
\label{puzzle:puzzle5}
\end{puzzle}
The clues of the puzzle appears in Listing~\ref{lst:zebrapuzzle2}.
For instance, \textit{"The green house is on the left of the white house."} is translated: $ green(x) \& white(y)  \rightarrow rightneighbor(y,x)$ (line 4 in Listing~\ref{lst:zebrapuzzle2}).
we used auxiliary predicates like \textit{differentFrom} or \textit{rightneighbor}.
To translate the sentence \textit{In each house lives a person with a different nationality} we used the symmetrical relation $differentFrom(x,y)  \rightarrow  differentFrom(y,x)$, and we implement the unique name assumption with statements like $differentFrom(a,b)$ for all individuals in the domain.
To define the relation between houses, we have: $rightneighbor(x,y) \vee rightneighbor(y,x) \leftrightarrow neighbor(x,y)$, which means that one is the neighbor of someone either if one lives just to his right or he lives just to your right. 
For \textit{"Each house has at least one nationality, pet, drink, color, car."}, we have sentences like 
$brit(x) \wedge swede(x) \wedge german(x) \wedge dane(x) \wedge norwegian$. 

\begin{lstlisting}[caption=A FOL theory used to provide Yes/No answers,linewidth=\columnwidth, firstline=19,lastline=33,breaklines=true,label=lst:zebrapuzzle2]
brit(x) <-> red(x).                       %the Brit lives in the red house
swede(x) <-> dog(x).                      %the Swede keeps dogs as pets
dane(x) <-> tea(x).                       %the Dane drinks tea
green(x) & white(y) -> rightneighbor(y,x). 
green(x) <-> coffee(x).
pallmall(x) <-> bird(x).
yellow(x) <-> dunhill(x).
milk(c).
norwegian(a).    
blend(x) & cat(y) -> neighbor(x,y).
horse(x) & dunhill(y) -> neighbor(x,y).
bluemaster(x) <-> beer(x).    
german(x) <-> prince(x).               
norwegian(x) & blue(y) -> neighbor(x,y). 
blend(x) & water(y) -> neighbor(x,y).
\end{lstlisting}

\begin{table}
\centering
\begin{small}
\begin{tabular}{l|l|l}\hline
Hypothesis & Theorem in FOL & Answer \\ \hline
German lives in house D. & $german(D)$ & Entailment  \\
The house B is blue. & $ blue(B) $ & Entailment  \\
The man in house A drinks water. & $water(A)$ & Entailment  \\
The man in house D owns the fish. &  $fish(D)$ & Entailment  \\\hline
The house A is blue.  & $blue(A)$ & Contradiction\\
Dane lives in house E. & $dane(E)$ & Contradiction\\
The man in house A drinks coffee. & $coffee(A)$ & Contradiction\\
The man in house C owns the horse. &  $horse(C)$ & Contradiction \\\hline
\end{tabular}
\end{small}
\caption{Question answering for Puzzle~\ref{puzzle:puzzle5}\label{tab:qa1_kk}}
\end{table}

The resulted theory contains 25 unary predicates: 5 unary predicates for nationality, 5 for colors, 5 for pets, 5 for drinks, and 5 for cigars, five houses: \textit{A, B, C, D,and E}. Hence, there are $5 \times 25 = 125$ questions. Among them 25 will generate entailments, and 100 will generate contradictions. A sample of these 8 questions are listed in Table~\ref{tab:qa1_kk}.
To extend the dataset, ambiguous puzzles are generated by removing clues. 
For instance, by removing the clue \textit{"The man living in the center house drinks milk"} (line 8 from Listing \ref{lst:zebrapuzzle2}), Mace4 generates 3 models. 
By removing  \textit{"The Norwegian lives in the first house"} (line 9), Mace4 generates 17 different models.\footnote{Moreover, the set of zebra puzzles can be easily increased by adding new zebra puzzle that can be generated automatically, e.g., ~http://new.mensus.net/brain/logic.shtml.}

\subsection{System Overview}

\begin{figure}
\begin{center}
 \resizebox{0.5\textwidth}{!}{%
\begin{tikzpicture}[node/.style={rectangle, draw=blue!50, fill=black!5, thick, minimum size=10mm},node distance=20mm,text width=5em,text centered, rounded corners]
\tikzstyle{level2} = [rectangle, draw, fill=blue!20, 
    text width=6em, text centered, rounded corners, minimum height=1em, node distance=4mm]
\node[rectangle, draw, fill=green!20, 
    text width=4em, text centered, rounded corners, minimum height=1em, node distance=4mm]      (a)       {Puzzles\\data set};
\node[node]      (b)   [right=of a]  {NL2FOL};
\node[node]      (c)   [right=of b]  {Generate Questions};
\node[node]      (d)   [below left=of c]  {Theorem Prover/ Model Finder };
\node[node]      (e)   [right=of d]  {Validate results};
\node[level2]    (f)   [above left=of b]  {grammar.fcfg};
\node[level2]    (g)   [right=of f]  {background.fol};
\node[level2]    (h)   [left=of d]  {mace4 background.in};

\draw[->] (a.east) -- (b.west) node[midway, above]{};
\draw[->] (b.east) -- (c.west) node[midway, above]{assumpt.txt};
\draw[->] (c.south) -- (d.north) node[midway, above]{questions.txt};
\draw[->] (b.south) -- (d.north) node[midway, above]{};
\draw[->] (d.east) -- (e.west) node[midway, above]{results.out};
\draw[->] (f.south) -- (b.north) node[near start, below]{};
\draw[->] (g.south) -- (b.north) node[near start, below]{};
\draw[->] (h.east) -- (d.west) node[near end, below]{};
\end{tikzpicture}
}%
\caption{Generating the dataset from natural language puzzles\label{fig:architecture}}
\end{center}
\end{figure}
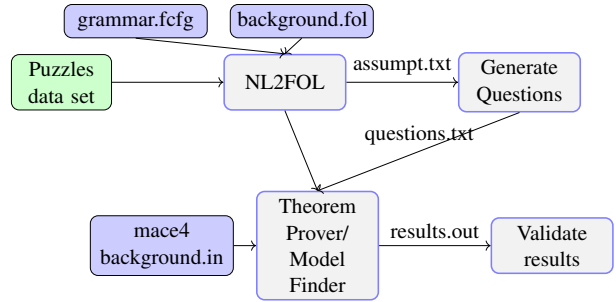

  The architecture for building the \puz\ dataset has four modules (Figure~\ref{fig:architecture}): (1) translating puzzles from natural language to FOL;
  (2) generating the entire set of questions for each puzzle;
  (3) computing answers by using theorem Mace4 model finder;
  (4) interpreting the results and generating the data set.

The first component takes each puzzle and translate line by line from natural language to FOL using the Natural Language Toolkit~(\newcite{perkins2014python}). 
The text is parsed using three grammars manually build for each puzzle domain: the fcfg grammar file (feature grammar file /context free grammar file), and a resource file with domain knowledge. 
Then, we generate the input file with assumptions for Mace4/Prover9 (e.g. Listings~\ref{lst:kk0bp} or  \ref{lst:kk0p}). 
The set of questions is generated automatically based on the assumptions in FOL, other existing predicates in the current puzzle.

Using the assumptions generated with the first module, and adding background knowledge file, we obtain the result for each question from questions list using Mace4/Prover9.
The background knowledge for comparison puzzles was built based on the transitivity (see Listing \ref{lst:bgk}), anti-symmetry, irreflexivity and trichotomy properties.

\begin{lstlisting}[caption=Adding background knowledge for comparison puzzles,linewidth=\columnwidth,breaklines=true,label=lst:bgk]
   (shorter(x,y) & shorter(y,z) -> shorter(x,z)).           %transitivity
   (taller(x,y)  & taller(y,z)  -> taller(x,z)).
   exists x (tallest(x) & (all y (tallest(y) -> y=x))).
   exists x (shortest(x) & (all y ((shortest(y) -> x=y)))).
   all x all y (taller(x,y) & taller(y,x) -> x=y).          %antisymmetry
   all x all y (shorter(x,y) & shorter(y,x) -> x=y).
  -taller(x,x & -shorter(x,x).                              %irreflexivity
   all x all y (shorter(x,y) | x=y | taller(x,y)).          %trichotomy
   (shorter(x,y) -> (x!=y & -taller(x,y))).
   (x=y) -> (-shorter(x,y) & -taller(x,y)).
   (taller(x,y) -> (x!=y & -shorter(x,y))).
   exists x (all y (-shorter(y,x)) & shortest(x)). 
   exists x (all y (-taller(y,x)) & tallest(x)). 
   taller(x,y) <-> shorter(y,x). 
\end{lstlisting}

Based on the result returned by Mace4 and Prover9~(\newcite{mccune2005prover9}), we compute the answer for each question. 
When Mace4 generates a single model, in case of unambiguous puzzle, the current question is entailed from puzzle. 
For ambiguous puzzles, Mace4 can generate multiple solutions - in this case we have the unknown relation.
The current version of the \puz\ dataset contains 16,745 pairs from which 4,467 are labeled with entailment, 8,530 with contradiction, and 3,748 unknown. 
\begin{table}
   \centering
   \begin{small}
    
     \begin{tabular}{|l|l|l|l|l||l|l|l|l|l|}\hline
    Puzzles & \multicolumn{4}{|l||}{Complete information} &  \multicolumn{5}{|l|}{Ambiguity (incomplete information)}   \\ \hline
     & No  & Questions & Ent. & Contr. & No & Questions & Ent. & Contr. & Unknown\\ \hline
        Comparison &\comppuzzlesamb & \withoutquestionsall & \withoutquestionsent & \withoutquestionscont &  \comppuzzleswamb &\withquestionsall & \withquestionsent & \withquestionscont & \withquestionsunknown \\ \hline
         Knights and knaves & \knightpuzzles & \knightquestionsall & \knightquestionsent & \knightquestionscont & \knightpuzzleswamb & \kwithquestionsall & \kwithquestionsent & \kwithquestionscont & \kwithquestionsunknown  \\ \hline
         Zebra & \zebrapuzzles & \zebraquestionsall & \zebraquestionsent & \zebraquestionscont & \ambzebrapuzzles & \ambzebraquestionsall & \ambzebraquestionsent & \ambzebraquestionscont & \ambzebraquestionsunknown  \\ \hline
         Total & \total & 4,136 & 1,475 & 2,661 & \totalamb & 16,745 & 4,467 & 8,530 & 3,748 \\ \hline
    \end{tabular}
   \end{small}
    \caption{Quantifying the PuzzTE dataset}
    \label{tab:puzzle}
\end{table}

\section{Discussion and related work}
By translating puzzles into FOL and than ask Mace4 for satisfiable models, our approach is also an automatic solution for solving puzzles. 
Solving logical puzzles is considered a challenging task, both for the human agent and the software agent. For the human agent, one can browse the 140 puzzles from the TPTP collection or the 144 puzzles modelled in FOL by \newcite{groza2021modelling}.
For the software agent, several puzzle solvers have been proposed~(\newcite{lev2004solving}, \newcite{milicevic2012puzzler}, \newcite{bogaerts2020framework}, \newcite{de2018predicate}, \newcite{jabrayilzade2020lgpsolver},~\newcite{mitra2015learning}, \newcite{groza2021natural}). 

\newcite{lev2004solving} have proposed a solution based on grammar rules, FOL, and model builders. 
Lev et al. have focused on puzzles of type "multiple-choice question", so the inferences have to just find out the correct answer, not to discover it. 
\newcite{milicevic2012puzzler} have tackled the task by using the ink Grammar general-purpose English parser, a semantic translator, and an automated logical analyzer. 
The solver is designed around the Zebra puzzles and tested on a dataset of 68 puzzles.
\newcite{bogaerts2020framework} have proposed a solver for logic grid puzzles which also makes use of QA and XAI. 
\newcite{de2018predicate} have developed the IDP system based on an extension of FOL.
A drawback is that the named entities (e.g. persons, colors) 
cannot be detected automatically and must be stated by the user.
Also, there exists a semi-automated process to detect the synonymy between verbs.
\newcite{jabrayilzade2020lgpsolver} have proposed the  DistilBERT tool that automatically solves logic grid  puzzles
The clues are translated in Prolog. 
The zebra puzzles have different categories that needs to be recognised within the text (e.g. person, name, occupation, color). 
\newcite{mitra2015learning} have developed the Logicia system also for 150 Zebra puzzles. 
The clues are classified using a maximum entropy model based on features like POS tags or dependency trees. 
The target language is answer set programming, based on which  71 out of 100 puzzles have been solved.
\newcite{groza2021natural} have also used grammar rules and named entity recognition to obtain a theory in FOL, which was given to Prover9 theorem prover. 
The grammar rules were manually created by analysing 43 puzzles for identifying the recurrent predicates, and then tested the resulted grammar on 331 puzzles. NER has been used to identify the characters in the puzzle (e.g., number of knights and knaves). A graphical representation of the proof is displayed in order to explain why an answer to the puzzles is correct or wrong.

\section{Conclusion}
We propose here the \puz\ dataset for textual entailment tasks. 
The three labels - entailment, contradiction and unknown - are automatically computed using model finders. This is the main distinguishing feature from the existing datasets, where labelling is performed by human experts or crowdsourcing. The \puz\ dataset exploits two properties of puzzles:
(i) each piece of information is necessary and (ii) no unnecessary information is provided. 
In our view, these properties make puzzles interesting candidates for machine comprehension tasks.

Ongoing work consists of organising a competition around this dataset on
Kaggle. 
The competition will include two tasks: (1) questions answering with complete information, and (2) questions answering with incomplete information. 
From the available pairs, 80\% will be available for training and validation, while the remaining 20\% for testing the submitted models.
We expect that machine learning approaches will face some difficulties, while approaches based on natural language understanding will be favored by our inference oriented dataset. 
However, both machine learning and symbolic reasoning approaches will benefit from the fact that the puzzles do not contain unnecessary information. 

The assure correct labels for each pair, the dataset was constructed in two steps.
First, the puzzles were automatically translated from text to FOL via grammar rules. 
Second, each translation was manually verified to identify possible translation errors. Given, the FOL theory, we use Mace4 to correctly computes the entailment, contradiction or ambiguity relations between the initial text and the answer. 
We are currently working on building a question-answering dataset without need of human checking. For this, we can employ a puzzle generator that builds puzzles directly in FOL. Then, we verbalise the FOL theory into natural language. Hence we will have both the FOL theory and the natural language puzzle.

\bibliographystyle{consilr}
\bibliography{consilr2021}

\begin{thebibliography}{}

\bibitem[\protect\citename{Bogaerts \bgroup et al.\egroup
  }2020]{bogaerts2020framework}
Bart Bogaerts, Emilio Gamba, and Tias Guns.
\newblock 2020.
\newblock A framework for step-wise explaining how to solve constraint
  satisfaction problems.
\newblock {\em arXiv preprint arXiv:2006.06343}.

\bibitem[\protect\citename{Bowman \bgroup et al.\egroup }2015]{bowman2015large}
Samuel~R. Bowman, Gabor Angeli, Christopher Potts, and Christopher~D. Manning.
\newblock 2015.
\newblock A large annotated corpus for learning natural language inference.

\bibitem[\protect\citename{Dagan \bgroup et al.\egroup }2005]{rte-n}
Ido Dagan, Oren Glickman, and Bernardo Magnini.
\newblock 2005.
\newblock The pascal recognising textual entailment challenge.
\newblock pages 177--190, 01.

\bibitem[\protect\citename{De~Cat \bgroup et al.\egroup }2018]{de2018predicate}
Broes De~Cat, Bart Bogaerts, Maurice Bruynooghe, Gerda Janssens, and Marc
  Denecker.
\newblock 2018.
\newblock Predicate logic as a modeling language: the idp system.
\newblock In {\em Declarative Logic Programming: Theory, Systems, and
  Applications}, pages 279--323.

\bibitem[\protect\citename{Groza and Nitu}2021]{groza2021natural}
Adrian Groza and Cristian Nitu.
\newblock 2021.
\newblock Natural language understanding for logical games.
\newblock {\em arXiv preprint arXiv:2110.00558}.

\bibitem[\protect\citename{Groza}2021]{groza2021modelling}
Adrian Groza.
\newblock 2021.
\newblock {\em Modelling Puzzles in First Order Logic}.
\newblock Springer.

\bibitem[\protect\citename{Hickl \bgroup et al.\egroup
  }2005]{Hickl05recognizingtextual}
Andrew Hickl, Jeremy Bensley, John Williams, Kirk Roberts, Bryan Rink, and Ying
  Shi.
\newblock 2005.
\newblock Recognizing textual entailment with lcc’s groundhog system.
\newblock In {\em In Proc. of the Second PASCAL Challenges Workshop}.

\bibitem[\protect\citename{Jabrayilzade and
  Tekir}2020]{jabrayilzade2020lgpsolver}
Elgun Jabrayilzade and Selma Tekir.
\newblock 2020.
\newblock Lgpsolver-solving logic grid puzzles automatically.
\newblock In {\em Proceedings of the 2020 Conference on Empirical Methods in
  Natural Language Processing: Findings}, pages 1118--1123.

\bibitem[\protect\citename{Khot \bgroup et al.\egroup
  }2018]{Khot_Sabharwal_Clark_2018}
Tushar Khot, Ashish Sabharwal, and Peter Clark.
\newblock 2018.
\newblock Scitail: A textual entailment dataset from science question
  answering.
\newblock {\em Proceedings of the AAAI Conference on Artificial Intelligence},
  32(1), Apr.

\bibitem[\protect\citename{Kotlerman \bgroup et al.\egroup
  }2015]{kotlerman-2015}
Lili Kotlerman, Ido Dagan, Bernardo Magnini, and Luisa Bentivogli.
\newblock 2015.
\newblock Textual entailment graphs.
\newblock {\em Natural Language Engineering}, -1:1--26, 06.

\bibitem[\protect\citename{Lev \bgroup et al.\egroup }2004]{lev2004solving}
Iddo Lev, Bill MacCartney, Christopher~D Manning, and Roger Levy.
\newblock 2004.
\newblock Solving logic puzzles: From robust processing to precise semantics.
\newblock In {\em Proceedings of the 2nd Workshop on Text Meaning and
  Interpretation}, pages 9--16.

\bibitem[\protect\citename{Magnini \bgroup et al.\egroup
  }2014]{magnini-etal-2014-excitement}
Bernardo Magnini, Roberto Zanoli, Ido Dagan, Kathrin Eichler, Guenter Neumann,
  Tae-Gil Noh, Sebastian Pado, Asher Stern, and Omer Levy.
\newblock 2014.
\newblock The excitement open platform for textual inferences.
\newblock In {\em Proceedings of 52nd Annual Meeting of the Association for
  Computational Linguistics: System Demonstrations}, pages 43--48, Baltimore,
  Maryland, June. Association for Computational Linguistics.

\bibitem[\protect\citename{Marelli \bgroup et al.\egroup
  }2014]{marelli-etal-2014-sick}
Marco Marelli, Stefano Menini, Marco Baroni, Luisa Bentivogli, Raffaella
  Bernardi, and Roberto Zamparelli.
\newblock 2014.
\newblock A {SICK} cure for the evaluation of compositional distributional
  semantic models.
\newblock In {\em Proceedings of the Ninth International Conference on Language
  Resources and Evaluation ({LREC}'14)}, pages 216--223, Reykjavik, Iceland,
  May. European Language Resources Association (ELRA).

\bibitem[\protect\citename{McCune}2005]{mccune2005prover9}
William McCune.
\newblock 2005.
\newblock Prover9.
\newblock {\em University of New M{\'e}xico}.

\bibitem[\protect\citename{Milicevic \bgroup et al.\egroup
  }2012]{milicevic2012puzzler}
Aleksandar Milicevic, Joseph~P Near, and Rishabh Singh.
\newblock 2012.
\newblock Puzzler: An automated logic puzzle solver.
\newblock {\em Dostopno na: http://people. csail. mit.
  edu/jnear/puzzler/writeup. html}.

\bibitem[\protect\citename{Minaee \bgroup et al.\egroup }2021]{minaee2021deep}
Shervin Minaee, Nal Kalchbrenner, Erik Cambria, Narjes Nikzad, Meysam
  Chenaghlu, and Jianfeng Gao.
\newblock 2021.
\newblock Deep learning based text classification: A comprehensive review.

\bibitem[\protect\citename{Mitra and Baral}2015]{mitra2015learning}
Arindam Mitra and Chitta Baral.
\newblock 2015.
\newblock Learning to automatically solve logic grid puzzles.
\newblock In {\em Proceedings of the 2015 Conference on Empirical Methods in
  Natural Language Processing}, pages 1023--1033.

\bibitem[\protect\citename{Perkins}2014]{perkins2014python}
Jacob Perkins.
\newblock 2014.
\newblock {\em Python 3 text processing with NLTK 3 cookbook}.
\newblock Packt Publishing Ltd.

\bibitem[\protect\citename{Rajpurkar \bgroup et al.\egroup
  }2016]{rajpurkar2016squad}
Pranav Rajpurkar, Jian Zhang, Konstantin Lopyrev, and Percy Liang.
\newblock 2016.
\newblock Squad: 100,000+ questions for machine comprehension of text.

\bibitem[\protect\citename{Thorne \bgroup et al.\egroup }2018]{thorne2018fever}
James Thorne, Andreas Vlachos, Christos Christodoulopoulos, and Arpit Mittal.
\newblock 2018.
\newblock Fever: a large-scale dataset for fact extraction and verification.

\bibitem[\protect\citename{Wagner}2010]{Wagner2010StevenBE}
Wiebke Wagner.
\newblock 2010.
\newblock Steven bird, ewan klein and edward loper: Natural language processing
  with python, analyzing text with the natural language toolkit.
\newblock {\em Language Resources and Evaluation}, 44:421--424.

\bibitem[\protect\citename{Williams \bgroup et al.\egroup
  }2018]{williams-etal-2018-broad}
Adina Williams, Nikita Nangia, and Samuel Bowman.
\newblock 2018.
\newblock A broad-coverage challenge corpus for sentence understanding through
  inference.
\newblock In {\em Proceedings of the 2018 Conference of the North {A}merican
  Chapter of the Association for Computational Linguistics: Human Language
  Technologies, Volume 1 (Long Papers)}, pages 1112--1122, New Orleans,
  Louisiana, June. Association for Computational Linguistics.

\bibitem[\protect\citename{Yang \bgroup et al.\egroup
  }2015]{yang-etal-2015-wikiqa}
Yi~Yang, Wen-tau Yih, and Christopher Meek.
\newblock 2015.
\newblock {W}iki{QA}: A challenge dataset for open-domain question answering.
\newblock In {\em Proceedings of the 2015 Conference on Empirical Methods in
  Natural Language Processing}, pages 2013--2018, Lisbon, Portugal, September.
  Association for Computational Linguistics.

\end{thebibliography}

\end{document}